\documentclass[sigconf,authorversion,nonacm]{acmart}

\copyrightyear{2025}

\usepackage{tabularx}

\begin{document}

\title{PrismAvatar: Real-time animated 3D neural head avatars on edge devices}

\author{Prashant Raina}
\authornote{Corresponding author.}
\email{prashantraina2005@gmail.com}
\orcid{0000-0002-1039-3472}
\affiliation{
  \institution{LG Electronics}
  \city{Toronto}
  \state{ON}
  \country{Canada}
}
\author{Felix Taubner}
\email{felix.taubner@lge.com}
\affiliation{
  \institution{LG Electronics}
  \city{Toronto}
  \state{ON}
  \country{Canada}
}
\author{Mathieu Tuli}
\email{mathieu.tuli@lge.com}
\affiliation{
  \institution{LG Electronics}
  \city{Toronto}
  \state{ON}
  \country{Canada}
}
\author{Eu Wern Teh}
\email{euwern.teh@lge.com}
\affiliation{
  \institution{LG Electronics}
  \city{Toronto}
  \state{ON}
  \country{Canada}
}
\author{Kevin Ferreira}
\email{kevin.ferreira@lge.com}
\affiliation{
  \institution{LG Electronics}
  \city{Toronto}
  \state{ON}
  \country{Canada}
}

\renewcommand{\shortauthors}{Raina et al.}

\begin{abstract}
    We present PrismAvatar: a 3D head avatar model which is designed specifically to enable real-time animation and rendering on resource-constrained edge devices, while still enjoying the benefits of neural volumetric rendering at training time. 
By integrating a rigged prism lattice with a 3D morphable head model, we use a hybrid rendering model to simultaneously reconstruct a mesh-based head and a deformable NeRF model for regions not represented by the 3DMM. 
We then distill the deformable NeRF into a rigged mesh and neural textures, which can be animated and rendered efficiently within the constraints of the traditional triangle rendering pipeline. 
In addition to running at 60 fps with low memory usage on mobile devices, we find that our trained models have comparable quality to state-of-the-art 3D avatar models on desktop devices.
\end{abstract}

\begin{teaserfigure}
  \includegraphics[width=\textwidth]{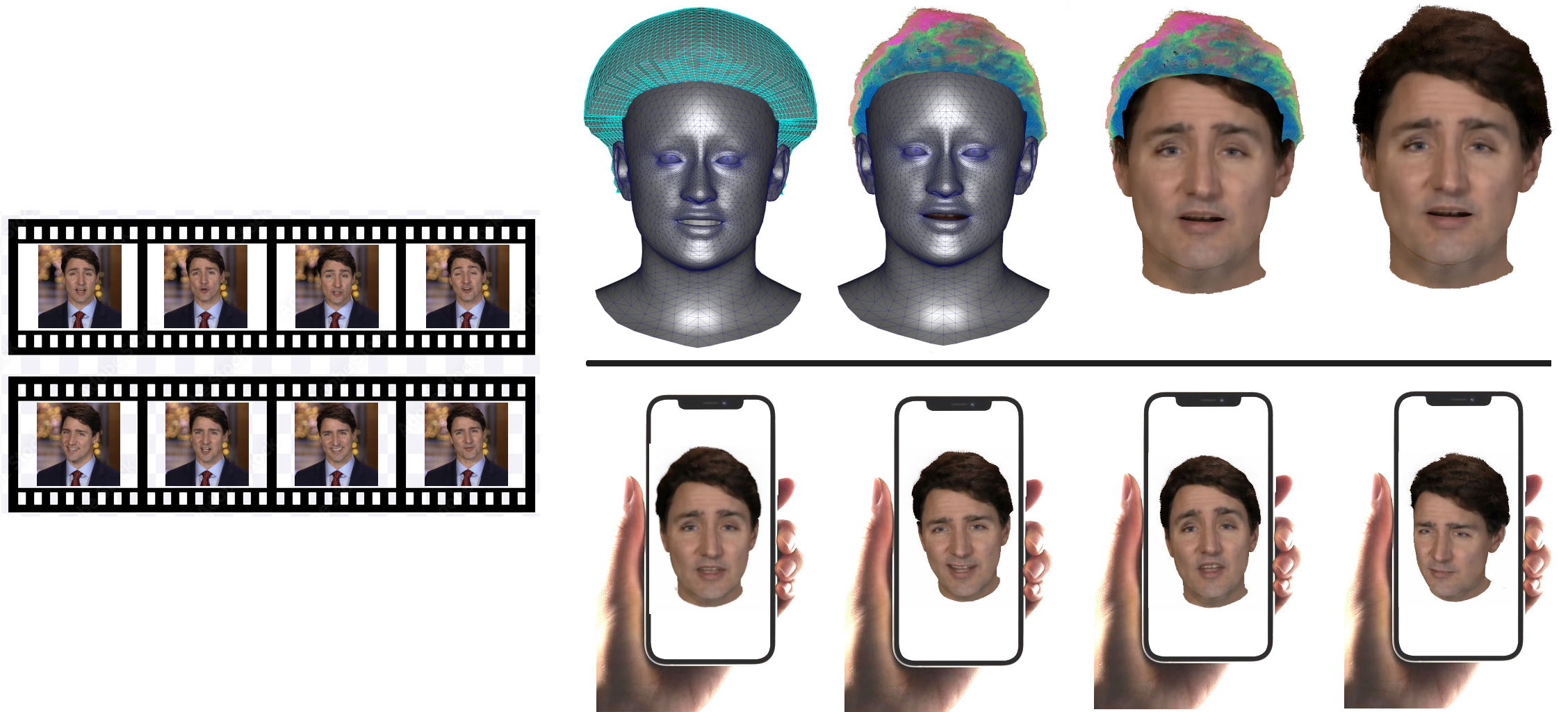}
  \caption{Our method trains a hybrid mesh-volumetric head avatar model from a given video, which we export to a rigged triangular mesh representation that allows real-time animation and rendering on resource-constrained edge devices.}
  \Description{Replace this with a text description of the teaser image for someone who cannot see it.}
  \label{fig_teaser}
\end{teaserfigure}

\maketitle

\section{Introduction}

Recent years have seen a surge of interest in photorealistic 3D human avatars, motivated by potential applications in teleconferencing, virtual assistants, entertainment and virtual reality experiences. Much of the latest research has focused on dynamic neural fields, such as deformable neural radiance fields (NeRFs). The main advantage offered by neural radiance fields is that they do not require explicit modeling of complex geometric details or the reflective properties of materials. However, even the more computationally efficient NeRF models generally demand a significant amount of GPU VRAM, as well as custom rendering pipelines based on general-purpose GPU libraries such as CUDA. This limits their usage in resource-constrained environments, such as mobile devices, smart TVs and car infotainment systems. It also restricts their use in environments with limited GPU programming capabilities, such as web applications. These drawbacks have limited their attractiveness for real-world product applications.

In this work, we present PrismAvatar, a novel model for dynamic 3D neural head avatars on edge devices. Our model maintains high rendering speed and compatibility with the constrained computing environment of mobile device browsers by using only the triangular mesh rendering pipeline of the GPU at inference time. At training time, however, we use a novel hybrid mesh-volume 3D representation in which the complex geometry of the hair is learned as a deformable neural radiance field. The deformation of the NeRF is controlled by a dense prism lattice which is rigged to move together with a 3D morphable model of the head. Grounding our model in a statistical 3DMM helps to ensure that our head avatars have plausible head shapes, while also allowing the model to be animated efficiently using a combination of blendshapes and linear blend skinning. It also allows us to leverage the strengths of mesh rendering for much of the face, thus avoiding NeRF artifacts such as floaters in the face region. For mobile device compatibility and efficient rendering at inference time, we take inspiration from recent work on static NeRF rendering for mobile devices~\cite{chen2023mobilenerf} to distill the canonical radiance field into a triangle soup with an associated neural texture. We implement animation and deferred neural rendering of these triangles using the basic vertex and fragment shading capabilities available on most  mobile devices and smart TVs.

In summary, our main contributions are:
\begin{enumerate}
    \item A technique for constructing a novel prism lattice structure that controls a deformable NeRF, allowing it to be animated by a 3D morphable model.
    \item A novel hybrid rendering approach to 3D head avatar reconstruction, which leverages the strengths of both surface and volume rendering. We demonstrate that this 3D avatar can be distilled into a purely mesh-based representation, allowing it to be rendered in real-time without significant loss of quality in constrained environments such as browsers on edge devices.
\end{enumerate}

\section{Related Work}

\subsection{3D Head Avatars}

Reconstructing 3D head avatars has traditionally involved multi-camera 3D reconstruction~\cite{bradley2010high,beeler2010high}, sometimes combined with markers or face paint~\cite{furukawa2009dense}. 
Most recent work has moved towards leveraging neural rendering to facilitate faster reconstruction from monocular videos, as well as to simplify the reconstruction of complex surfaces and materials. 
Implicit neural representations of surfaces such as signed distance functions SDFs~\cite{zheng2022avatar} and point clouds~\cite{zheng2023pointavatar, wang2023npbva} have seen a lot of use due to their ability to smoothly change topology during the reconstruction process. 
In addition, volumetric neural rendering approaches have been explored due to the potential for more photorealistic reconstruction of hair and other regions of the head with complex geometry~\cite{lombardi2019neural, lombardi2021mvp, cao2022ava}.
3D morphable models (3DMMs) of heads such as the Basel Face Model~\cite{blanz1999bfm} and FLAME~\cite{li2017flame} are often incorporated into models to guide the deformation of an implicit field. 
Such approaches range from simply conditioning a volumetric field on blendshape coefficients~\cite{gafni2021nerface}, to treating a 3DMM as a proxy surface model which guides the deformation of a volumetric field~\cite{athar2022rignerf,athar2023flameinnerf,bai2023monoavatar,zielonka2023insta}. 
Another line of work focuses on learning latent representations of 3D avatars instead of fields or explicit shapes~\cite{xu2023latentavatar,ma2021pixel}. 
Such methods would require real-time neural network inference if deployed on edge devices, which places them well out of reach of cross-platform edge device browsers with minimal GPU programming capabilities.
There are also methods such as Neural Head Avatars (NHA)~\cite{grassal2022nha}, ROME~\cite{khakhulin2022rome} and FLARE~\cite{bharadwaj2023flare}, which use triangular meshes as the underlying representation.
Our final exported model is also a triangular mesh representation with a neural texture~\cite{thies2019deferred}. However, we use a hybrid mesh-volumetric model at training time.

\begin{figure*}[h!]
    \includegraphics[width=250pt]{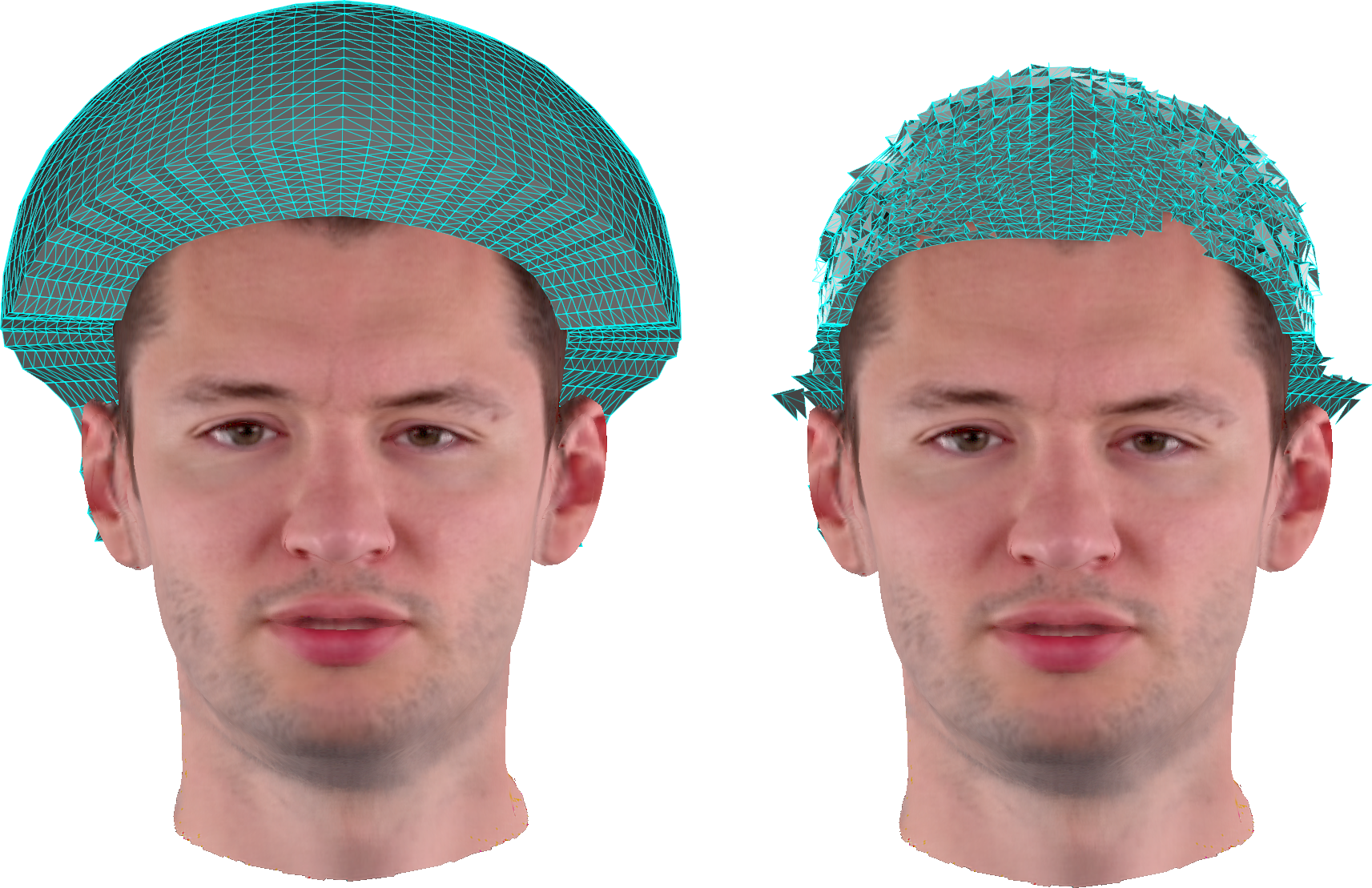}
    \caption{Left: The volumetric field for hair in our hybrid model is defined over a prism lattice, constructed as described in Section \ref{subsec_lattice}. Right: At the time of export (Section \ref{subsec_model_export}), we prune the lattice to remove triangles that are invisible or occluded.}
    \Description{Replace this with a text description of the figure for someone who cannot see it.}
    \label{fig_lattices}
\end{figure*}

\subsection{Fast NeRF Rendering}

Since the inception of NeRFs~\cite{mildenhall2021nerf}, numerous methods have been proposed to improve the speed of training and inference. 
Several methods use regular grids with precomputed features to improve rendering speed. FastNeRF~\cite{garbin2021fastnerf} and Plenoxels~\cite{fridovich2022plenoxels} both use dense voxel grids, although they differ in the nature of the features stored in the grid. KiloNeRF~\cite{reiser2021kilonerf} uses a grid of thousands of small MLPs that specialize in reconstructing a small portion of a scene. PlenOctrees~\cite{yu2021plenoctrees} uses a sparse voxel octree containing opacity and color represented as spherical harmonic coefficients (as is the case with Plenoxels).
SNeRG~\cite{hedman2021snerg} ``bakes'' a trained NeRF into a grid data structure suitable for fast rendering, and is the first NeRF implementation that was shown to run in real-time on low-resource mobile devices.
The aforementioned methods obtain their speedups at the expense of high memory requirements to store data structures and associated features.

AutoInt~\cite{lindell2021autoint} learns partial volume rendering integrals along each ray in order to reduce network evaluations by piecewise ray-marching. R2L~\cite{wang2022r2l} later showed that it is possible to learn neural light fields (NeLFs) by distilling NeRFs. This approach can be considered as equivalent to learning complete integrals along each ray in a NeRF. DyLin~\cite{yu2023dylin} extended R2L to dynamic scenes by conditioning on either time or arbitrary control parameters. MobileR2L~\cite{cao2023mobiler2l} and LightSpeedR2L~\cite{gupta2023lightspeed} later showed that the R2L distillation approach can be used to produce light fields that can be evaluated efficiently on mobile devices. 

In contrast to the light field distillation methods, MobileNeRF~\cite{chen2023mobilenerf} showed that a NeRF trained on a domain covered by a regular grid can be distilled into a set of triangles that can be efficiently rendered with a neural texture on mobile devices. 
Concurrently with our work, adaptive shells~\cite{wang2023adaptive} and VMesh~\cite{guo2023vmesh} have sought to obtain speedups via hybrid mesh-volume reconstruction of static scenes, with the hybrid NeRF-SDF model NeuS~\cite{wang2021neus} as their starting point.
We take a similar approach to MobileNeRF, given our focus on edge device compatibility. Unlike MobileNeRF, which uses a regular cubic grid and can only handle static scenes, we reconstruct an animatable head avatar model.

\section{Method}
\label{sec_method}

\subsection{Overview}
\label{subsec_method_overview}

Our method takes as input a series of matted images of a head, along with camera and 3DMM parameters obtained using a head tracker. These images are used to train a hybrid mesh-volumetric model of a head using an analysis-by-synthesis approach. The mesh representation is based on the FLAME 3D morphable model~\cite{li2017flame}, which adequately models much of the face and neck. The 3DMM does not model hair, and hair is not easily modeled by a surface representation such as a mesh. Therefore, we use a neural radiance field (NeRF)~\cite{mildenhall2021nerf} as the volumetric representation for scalp hair and facial hair. In order to allow the NeRF to move and deform along with the motion of the head, we construct a rigged prism lattice structure on top of the 3DMM mesh (Section \ref{subsec_lattice}). Ray intersections with this lattice are mapped to the canonical space of the undeformed model. The NeRF is represented by three networks (described in Section \ref{subsec_model_networks}), which allows sampling of a view-indepdendent neural feature at points in the canonical space. We use a hybrid rendering approach (Section \ref{subsec_model_rendering}) to train the NeRF in two stages. Finally, we export a compact triangle-based avatar model from the hybrid model (Section \ref{subsec_model_export}), which allows for efficient animation and neural surface rendering on edge devices.

\subsection{Prism Lattice}
\label{subsec_lattice}

Figure \ref{fig_lattices} shows an example of a prism lattice constructed on a FLAME mesh.
Starting with the FLAME template mesh, we manually mark regions of the mesh to be used as the base of prism lattices.  
For example, to reconstruct hair on the scalp, we mark triangles belonging to the scalp. We then subdivide each of these triangles into 4 or 16 triangles to ensure a sufficiently dense lattice. The subdivided triangles are then extruded along the normal of each vertex, forming a layer of prisms. These extrusions are repeated to obtain a stack of up to 16 layers of prisms. We repeat this extrusion process separately for all 400 FLAME shape and expression blendshapes. This allows us to create a new blendshape model that incorporates the prism lattice as part of the 3DMM. The extruded vertices share the linear blend skinning weights and pose corrective blendshapes of the base vertices, thereby allowing them to respond to skeletal animation of the FLAME model. To reconstruct facial hair, we add additional lattices for the sideburns and around the mouth (shown in Figure \ref{fig_results_beard}).

\subsection{Trained Networks}
\label{subsec_model_networks}

\begin{figure}
    \includegraphics[width=250pt]{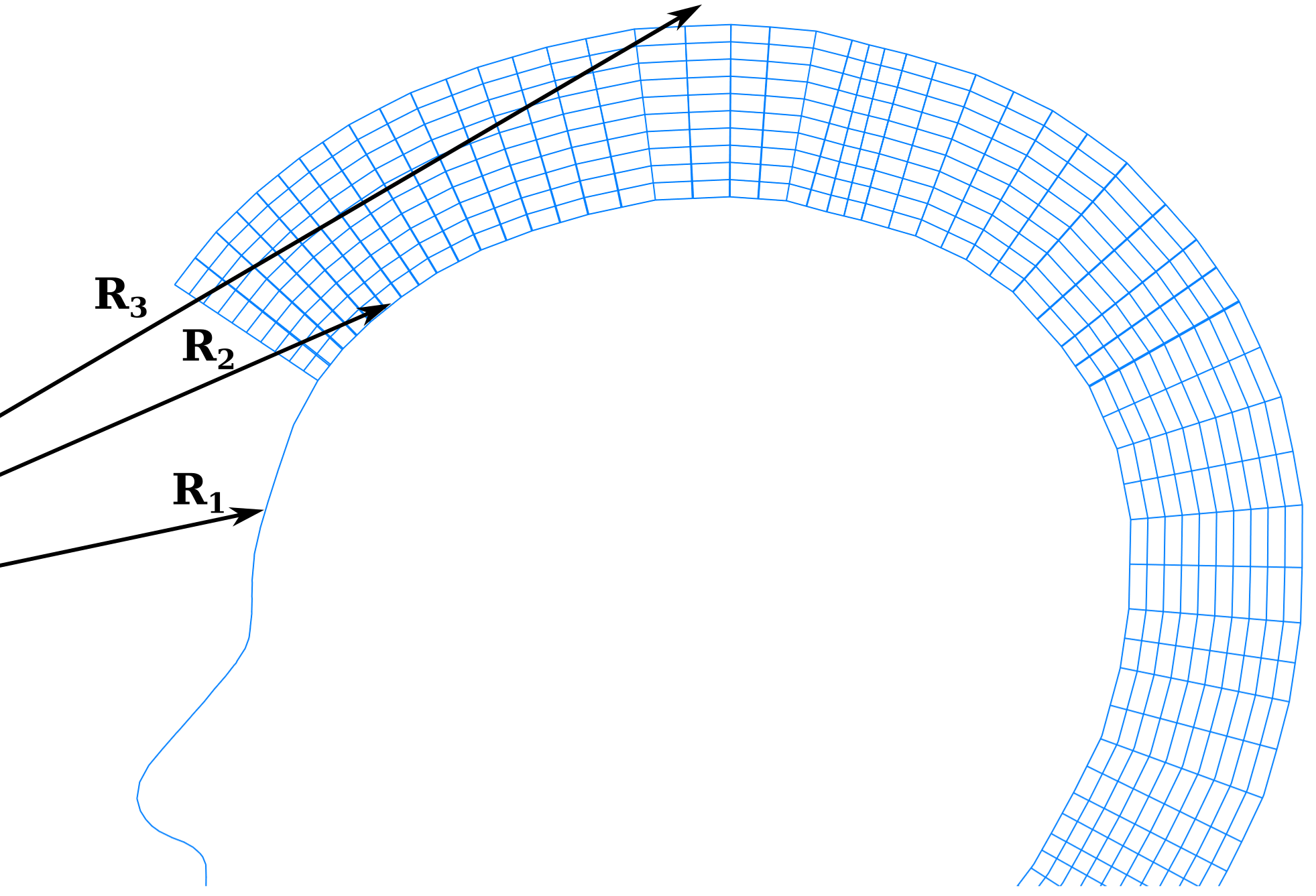}
    \caption{Illustration of different scenarios in ray intersection. Ray $R_1$ hits the FLAME mesh first, so its color is sampled from a learned texture. Ray $R_3$ intersects only triangles of the lattice, so its color is obtained by a volume rendering integral over the intersection points. Ray $R_2$ intersects the lattice before terminating on the FLAME mesh, so its color is obtained by interpolating the volume rendering integral before the last intersection with the color of the final intersection, using the accumulated opacity as the interpolation factor.}
    \Description{Replace this with a text description of the figure for someone who cannot see it.}
    \label{fig_hybrid_rays}
\end{figure}

\begin{figure*}[h!]
    \centering \includegraphics[width=\textwidth]{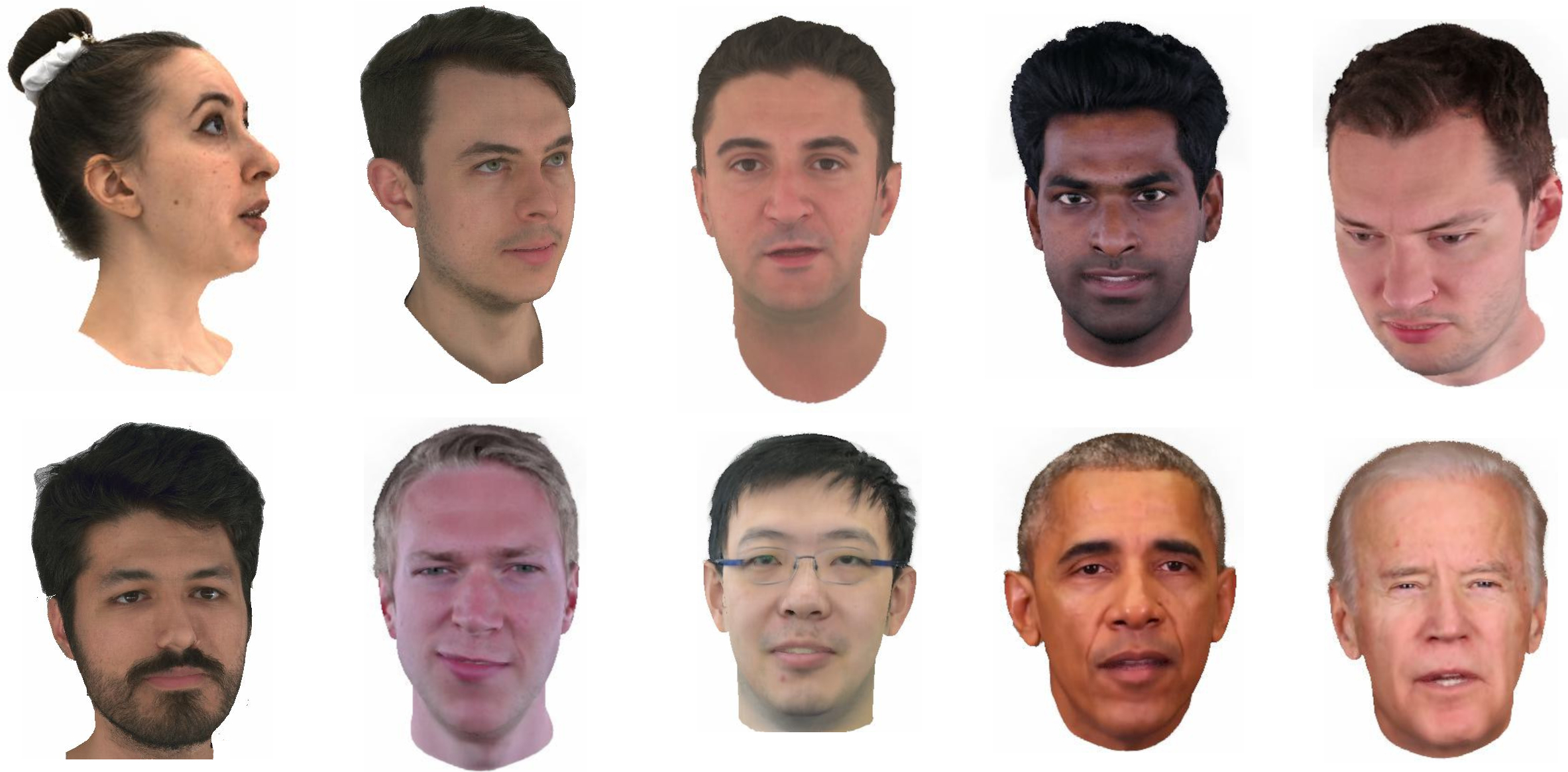}
    \caption{A sample of head avatars reconstructed using our method.}
    \Description{Replace this with a text description of the figure for someone who cannot see it.}
    \label{fig_results_collage}
\end{figure*}

Our NeRF network architecture differs in a couple of ways from traditional NeRF models. Firstly, we predict opacity or \textit{alpha} for sample points when integrating along rays, similar to previous work~\cite{lombardi2019neural, attal2022learning, chen2023mobilenerf}. Using alpha instead of the density which is used in traditional NeRF models enables our final exported model to utilize alpha testing during rendering. Secondly, we compute the view-dependent color at a point in two steps, similar to MobileNeRF~\cite{chen2023mobilenerf}. The first step produces a view-independent neural feature vector. The second step produces a color conditioned on the view direction.
We train three neural networks for the volumetric regions:
\begin{enumerate}
    \item An opacity prediction network $\mathcal{A}$, which predicts the opacity associated with a point in the canonical space of the NeRF. Predictions from the final trained network are used to generate alpha textures for the exported model. For a canonical point $\textbf{p}_k$, the predicted opacity is given by:
    \begin{flalign}
        && \alpha_k = \mathcal{A}(\textbf{p}_k;\theta_\mathcal{A})
        && \mathcal{A}:\mathbb{R}^3 \rightarrow [0,1]
    \end{flalign}
    \item A feature prediction network $\mathcal{F}$, which predicts an 8D neural feature vector associated with a point in the canonical space of the NeRF. Predictions from the final trained network are used to generate a neural texture for the exported model. For a canonical point $\textbf{p}_k$, the predicted feature is given by:
    \begin{flalign}
        && \textbf{f}_k = \mathcal{F}(\textbf{p}_k;\theta_\mathcal{F})
        && \mathcal{F}:\mathbb{R}^3 \rightarrow [0,1]^8
    \end{flalign}
    \item A small color prediction network $\mathcal{C}$, which predicts a color given a viewing direction and an 8D neural feature. The weights of this network are exported along with the avatar and used for deferred neural rendering on the edge device. For a feature vector $\textbf{f}_k$, and ray direction $\textbf{d}$, the predicted color is given by:
    \begin{flalign}
        && c_k = \mathcal{C}(\textbf{f}_k, \textbf{d};\theta_\mathcal{C})
        && \mathcal{C}: [0,1]^8 \times [-1,1]^3 \rightarrow [0,1]^3
    \end{flalign}
\end{enumerate}
The color prediction network is a simple multilayer perceptron (MLP) which concatenates an 8D neural feature vector with a 3D direction and passes them through two hidden layers of 16 neurons, with ReLU activations and sigmoid activation at the end to obtain a linear RGB color. Both the opacity prediction network and the feature prediction network utilize a learned hash-grid positional encoding and fully fused MLP architecture based on InstantNGP~\cite{muller2022instantngp}. This allows them to maximize their capacity for portions of the unit cube which are occupied by volumetric data. The two fully fused MLPs have 4 hidden layers of 128 neurons, with ReLU activations in between and sigmoid activation at the end.

\subsection{Hybrid Rendering}
\label{subsec_model_rendering}

Our model is rendered in a hybrid manner in which the areas enclosed by the prism lattices are rendered as deformable NeRFs, with the texture of the FLAME mesh incorporated into the ray integration in semi-transparent regions. By contrast, the base FLAME mesh is treated as an ordinary textured mesh with a learned texture and a learned alpha map. To handle the mouth interior, we fill the hole in the FLAME mesh for the mouth cavity. The triangles of the sealed hole are rendered with a learned 2D texture for the mouth cavity.

In standard NeRF rendering, ray marching is used to obtain sample points for integration along each ray. We do not take this approach. Instead, we exploit the efficiency of GPU raytracing hardware to intersect the ray with the lattice triangles. This provides us with a dense set of sample points along rays that pass through the volumetric portion of the model.
We use a custom differentiable renderer that shoots rays which intersect with both the FLAME mesh as well as the prism lattices. Rays either stop as soon as they hit the FLAME mesh, or continue to collect intersections with the prism lattices, up to a maximum of 64 intersections (illustration in Figure \ref{fig_hybrid_rays}). A triangle bounding volume hierarchy (BVH) is used to accelerate ray intersection tests.

When a ray hits the base FLAME mesh, the intersection point is assigned an opacity of $1$, and its color is sampled from the learned face texture instead of the neural rendering pipeline. Rays that hit the mouth cavity triangles are similarly treated as opaque, but with a separate learned texture.

There are two stages of training. The first stage trains all three networks, while the second stage freezes the opacity network. The equations for the integrated pixel intensity over intersections $\textbf{p}_k$ along a ray with origin $\textbf{o}$ and direction $\textbf{d}$ are:
\begin{equation}
    \textbf{I}(\textbf{o}, \textbf{d}) = \sum_{k=1}^{K} T_k \alpha_k \mathcal{C}(\mathcal{F}(\textbf{p}_k), \textbf{d}) 
\end{equation}
\begin{equation}
    T_k = \prod_{l=1}^{k-1}(1 - \alpha_l)
\end{equation}
Here, $\alpha_k$ is either computed using the opacity network for ray-lattice intersections, or is assigned a fixed opacity of $1$ in the case of ray-FLAME intersections. The $\mathcal{C}$ term is replaced with a color sampled from the learned face texture for ray-FLAME intersections.

The second stage of training changes the rendering algorithm to mimic the effects of triangle rasterization and deferred neural rendering. In this stage, each ray has exactly one associated feature vector, which is obtained as a weighted average of predicted feature vectors along the ray. The color network $\mathcal{C}$ is therefore only executed once per pixel. The ray integration for the second stage is given by:

\begin{equation}
    \label{eq_stage2_rendering}
    \textbf{I}(\textbf{o}, \textbf{d}) = \mathcal{C}\left(\sum_{k=1}^{K} T_k \alpha_k \mathcal{F}(\textbf{p}_k), \textbf{d}\right)
\end{equation}

This modified formulation also requires a different approach to handling rays that intersect both the lattice and the FLAME mesh. If the final ray intersection $K$ hits the FLAME mesh, we first compute the intensity excluding that intersection as $\textbf{I}_{lat}$. We sample the color of the FLAME mesh $\textbf{I}_{flame}$ at the intersection point from the learned texture, and then linearly interpolate between the two colors based on the accumulated transmittance up to that point:
\begin{equation}
    \textbf{I} = (1 - T_K)\textbf{I}_{lat} + T_K \textbf{I}_{flame}
\end{equation}

\subsection{Training Losses}
\label{subsec_model_losses}

There are two losses which are minimized in both training stages:
\begin{enumerate}
    \item A photometric loss $L_{photo}$, which is an $\ell_1$ log-sRGB loss used in previous work~\cite{hasselgren2021nvdiffmodeling, munkberg2022nvdiffrec}.
    \item An alpha or silhouette regularizer $L_{alpha}$, which is the mean square of the predicted alpha $\sum_{k=1}^{K} T_k \alpha_k$ for all pixels corresponding to the ground truth background mask. This encourages the networks to explain unoccupied regions by transparency as opposed to opaque surfaces with the background color.
\end{enumerate}

In addition, in the second stage we wish to binarize the predicted alpha values so that they can later be used for alpha testing (as opposed to alpha blending). This is done by applying a straight-through estimator, in line with previous work~\cite{bengio2013estimating, chen2023mobilenerf}. However, in order to better stabilize training, we calculate two versions of all losses: with and without the straight-through estimator. We gradually interpolate from the non-binarized losses to the binarized losses over the course of training.

\subsection{Model Export}
\label{subsec_model_export}

\begin{figure*}[h!]
    \centering \includegraphics[width=\textwidth]{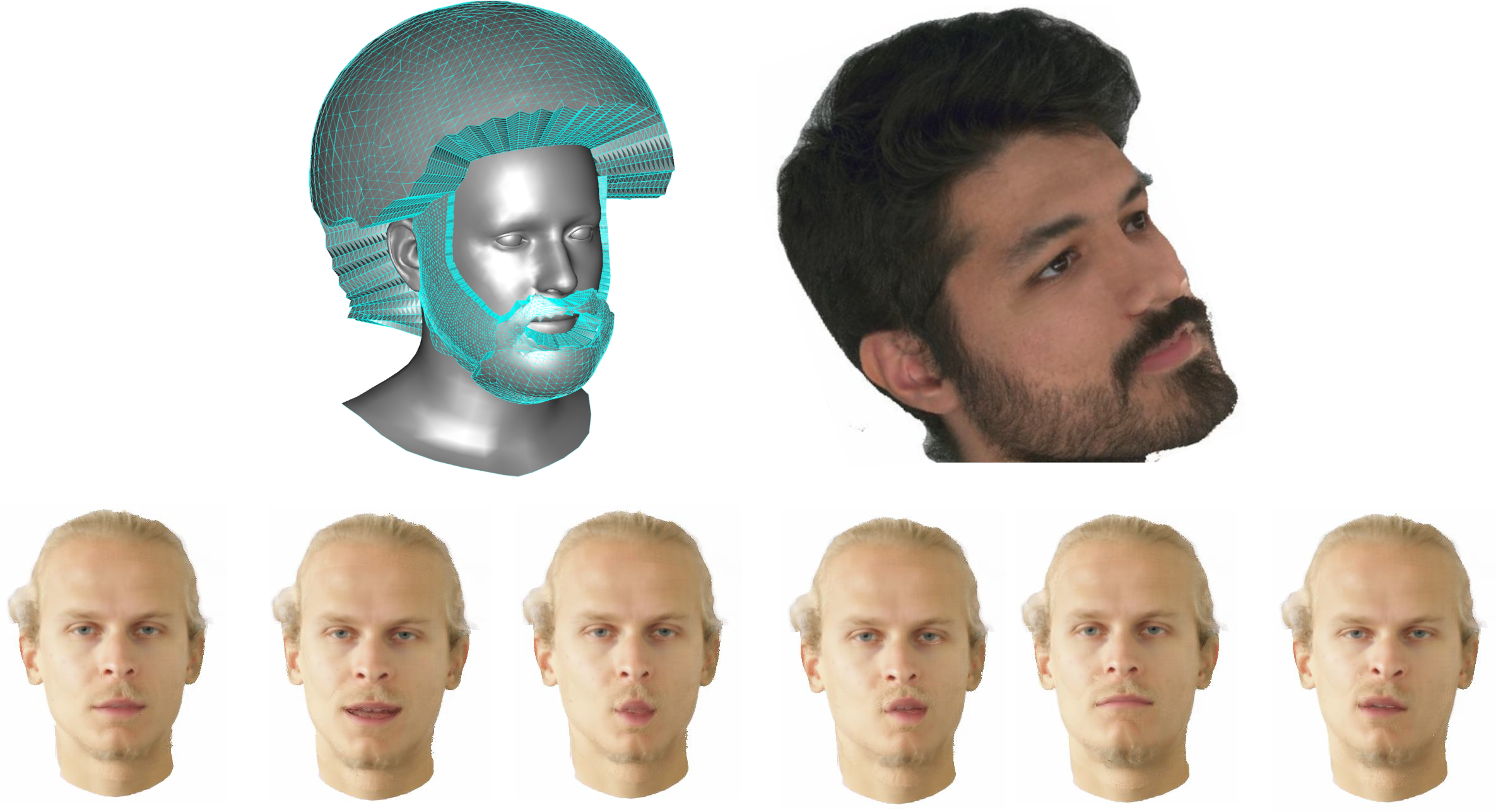}
    \caption{Using a prism lattice which covers the face allows us to reconstruct facial hair. Thick dark hair and thin blond hair are both reconstructed by our method. Top: The prism lattice for facial hair and an example of a reconstructed avatar. Bottom: Frames showing the deformation of the mustache in response to changes in the facial expression.}
    \Description{Replace this with a text description of the figure for someone who cannot see it.}
    \label{fig_results_beard}
\end{figure*}

The model we wish to export is a rigged triangular mesh with accompanying texture maps for combining deferred neural rendering with standard texture rendering. In order to obtain this triangular mesh, we export the entire FLAME mesh, but we cannot simply export the entire prism lattice. This is because the majority of the triangles in the lattice would be either completely transparent or occluded by other triangles. Therefore, the first export step is to prune occluded and transparent triangles. For each view in the training frames, we shoot rays through every pixel towards the prism lattice. The alpha values at the hit points are obtained by transforming them to the canonical space and passing them through the opacity network. If the predicted opacity is less than $0.5$, the ray continues on its trajectory, otherwise we stop the ray and record which triangle was hit. Repeating this process for different views gives us a list of triangles to keep. We then re-index the prism lattice without the pruned triangles.

There are three large texture maps which we generate for deferred neural rendering of the lattice triangles: an alpha map and two feature maps. Having two feature maps allows us to split the 8D feature vectors into two sets of RGBA channels, enabling compression of the textures as PNG. Each lattice triangle is assigned to a $16\times 16$ square cell of each texture map. The texture maps can be regarded as grids of these square cells, where the height and width of the grids are obtained by taking the square root of the number of lattice triangles and rounding up. For each square cell, we uniformly sample a $16\times 16$ grid of 3D points on the corresponding canonical lattice triangle. Each of the sampled 3D points is passed through the opacity and feature networks, giving us the values to be inserted into the alpha and feature maps. 
The uniform sampling of points over the triangle is achieved using a low-distortion mapping from a square to a triangle~\cite{heitz2019low}, allowing us to utilize the entire square cell of the texture map to store texture data for a triangle. The tradeoff of this mapping is that for easy rendering at inference time, we need to split every triangle in two, in order to have UV coordinates pointing to all four corners of the corresponding square region of the texture map. 

Both the FLAME mesh as well as our UV mapping of the prism lattice result in multiple UV coordinates corresponding to the same triangle vertex. For compatibility with traditional rendering pipelines, we duplicate such vertices so that each vertex has a unique UV coordinate. Besides the rigged mesh and textures, we also export the weights of the color network $\mathcal{C}$, to be used in the neural shader.

\section{Implementation}
\label{sec_implementation}

\begin{table*}[h!]
    \centering

    \begin{tabularx}{300 pt}{X|c|c|c|c}
    \hline
    Method & PSNR$\uparrow$          & SSIM$\uparrow$           & MS-SSIM$\uparrow$        & LPIPS$\downarrow$           \\ \hline
    PointAvatar~\cite{zheng2023pointavatar}   & 25.0 & 0.903 & 0.936 & 0.0717 \\ 
    FLARE~\cite{bharadwaj2023flare}   & 27.9 & 0.904 & 0.946 & 0.0602 \\ 
    INSTA~\cite{zielonka2023insta}   & \textbf{32.5} & \textbf{0.953} & \textbf{0.977} & \textbf{0.0453} \\ \hline
    Ours (before binarization)   & 31.3 & 0.942 & 0.970 & 0.0590 \\ 
    Ours (after binarization)   & 32.0 & 0.944 & 0.973 & 0.0593 \\ 
    Ours (after export)   & 30.4 & 0.929 & 0.960 & 0.0690 \\ 
    \hline
    \end{tabularx}
    \caption{Quantitative evaluation }
    \label{table_quantitative}
\end{table*}

We implemented separately the pipeline for training our \mbox{PrismAvatar} models, and a viewer web app compatible with mobile device browsers. We also implemented a headless renderer that reuses the rendering code of the viewer web app to allow accurate quantitative evaluation of images rendered from our exported models.

\subsection{Training}
Our training pipeline is implemented in PyTorch, with certain components implemented as custom Python modules written in C++. Our raytracing module was implemented using NVIDIA Optix 8.0, in order to utilize RTX cores to accelerate intersection tests. Using RTX cores for acceleration is made possible by the fact that we are intersecting rays with triangles, and not with a volume or implicit surface. We rebuild the BVH acceleration structure from scratch at the start of each training epoch to avoid degradation of the acceleration structure due to frequent updates.

Each training batch is a random sample of $2^{16}$ rays for each training image. We trained our models on a single NVIDIA A6000 GPU. The models are trained on 90,000 batches for each training stage, taking around 90 minutes per stage on average. We use the Adam optimizer~\cite{kingma2015adam}, with a learning rate of $5\times 10^{-5}$ for neural networks and $5\times 10^{-4}$ for learned textures.

\subsection{Viewer Web App}
In order to maximize cross-platform compatiblity across different edge devices for the 3D avatar viewer while also having fast performance, we implemented our viewer as a static web application using C++ compiled to WebAssembly via Emscripten~\cite{emscripten}. Our exported model contains a single rigged mesh, with different shading for the FLAME head and the triangle soup extracted from the prism lattice. The viewer renders the model in two passes:
\begin{enumerate}
    \item The FLAME mesh is skinned and shaded with the learned texture and alpha map. Then, the lattice triangles are skinned and shaded using a fragment shader implementation of the color network $\mathcal{C}$.
    \item A post-processing pass converts the linear RGB color to sRGB.
\end{enumerate}

\subsection{Video Preprocessing}
Our method requires background matting for input images, as well as camera information and 3DMM parameters for each frame. In order to conserve memory, high-resolution videos are downsampled to have a width or height of 512 pixels. \\
\textbf{Background removal.} The backgrounds of videos are removed using Robust Video Matting~\cite{lin2022robust} if no background mask is provided. This is followed by the removal of regions classified as clothing by a pretrained BiSeNet~\cite{yu2018bisenet}. \\
\textbf{Head tracking.} The videos are processed using a head tracker to provide camera intrinsics, camera extrinsics and FLAME 2023 model parameters for every frame. We use FlowFace~\cite{flowface} for head tracking on both monocular as well as multi-view videos.
\section{Results}

We have tested our method on monocular videos released with INSTA~\cite{zielonka2023insta}, as well as multi-view videos released with NerSemble~\cite{kirschstein2023nersemble} and the RenderMe-360 dataset~\cite{pan2023renderme}. Some of the reconstructed avatars are shown in Figure \ref{fig_results_collage}. We can see that hair is accurately reconstructed along with the face. Figure \ref{fig_results_beard} shows additional results highlighting our method's performance on facial hair. The prism lattice covering the face can not only be used to accurately reconstruct facial hair, but also to deform the hair in response to different facial expressions.

The trained and exported models were animated using head tracking data in our viewer web app. They were verified to run at 60 fps on iPhone Pro 14, as well as a 4th generation iPad Pro. They are also fully compatible with the Samsung Galaxy S9, an Android phone released in 2018. The framerate appears to be capped using VSync, since it never fluctuates above or below 60 fps. We were unable to disable VSync due to the implementation constraints imposed by the web browser. Therefore, the uncapped framerate could potentially be higher than 60 fps.

The average download size for the avatar is 70 MB. Google Chrome tabs running the avatar viewer use 206 MB CPU RAM and 46 MB GPU VRAM on average. The low VRAM usage reflects the simplicity and efficiency of our hybrid representation of the head.

We also measured the quality of our rendered images at different stages of the avatar generation process (Table \ref{table_quantitative}). We find that the opacity binarization during the second stage of training suppresses small artifacts in the model, resulting in a slight improvement in the metrics. As to be expected, there is a slight drop in the image quality introduced by the model export process. The metrics are compared against three recent head avatar reconstruction methods: PointAvatar~\cite{zheng2023pointavatar}, FLARE~\cite{bharadwaj2023flare} and INSTA~\cite{zielonka2023insta}. Since the publicly released implementations of these methods use monocular head tracking, we only use the monocular videos released with INSTA for a fair quantitative comparison. We use the original non-overlapping train-test splits for each video. The results show that, despite the fact that our method prioritizes edge device compatibility and produces a compact distilled model, our image quality metrics are still competitive with current state-of-the-art avatar methods which run on desktop devices. 

\section{Conclusion}

In this paper, we have presented a hybrid mesh-volumetric model for 3D head avatars which is specially constructed with a rigged prism lattice. We have shown that our model reconstructs the geometry and texture of hair on both the scalp and the face. It can also be exported into a triangle-based model which can be animated at real-time frame rates on edge devices with image quality comparable to the state of the art. We have validated our exported models on mobile devices with a cross-platform implementation. Being a web application, we believe our models have the potential to run on other edge devices such as smart TVs and car infotainment systems, as long as a suitable web browser with WebGL support is available.

\textbf{Limitations and Future Work.}
While our current method can reconstruct hair, our approach cannot model physics-based animation such as the effect of gravity on hair. This is a limitation shared by previous work that does not explicitly model individual strands or strips of hair. Our model of the mouth interior is also quite simple, and does not deal with complex motions such as protruding the tongue. In future work, we intend to handle these cases. We also intend to incorporate a physically-based lighting model to allow for the separation of lighting and material properties during avatar reconstruction.

\bibliographystyle{ACM-Reference-Format}
\bibliography{refs}

\end{document}